%% file: main.tex
\documentclass{article}

\PassOptionsToPackage{numbers, compress}{natbib}
\usepackage[preprint]{neurips_2024}

\usepackage{times}
\usepackage{latexsym}
\usepackage{inconsolata}

\usepackage[utf8]{inputenc} 
\usepackage[T1]{fontenc}    
\usepackage{hyperref}       
\usepackage{url}            
\usepackage{booktabs}       
\usepackage{amsfonts}       
\usepackage{nicefrac}       
\usepackage{microtype}      
\usepackage{xcolor}         

\usepackage{multirow}
\usepackage{subfigure}

\usepackage[english]{babel}
\usepackage{moresize}
\usepackage{amsmath}
\usepackage{algorithmic}
\usepackage{balance}
\usepackage{comment}
\usepackage{paralist}
\usepackage{bm}
\usepackage{pgfplots}
\usetikzlibrary{pgfplots.dateplot}

\usepackage{flushend}
\usepackage[english]{babel}
\usepackage{graphicx}

\usepackage{amssymb}
\usepackage{url}
\usepackage{bbm}
\usepackage{longtable}
\usepackage{rotating}
\usepackage{mathrsfs}
\usepackage{enumitem}
\usepackage[linesnumbered,algoruled,boxed,lined]{algorithm2e}
\usepackage{adjustbox}
\usepackage{hyperref}
\usepackage{filecontents}

\usepackage{tikz}
\usepackage{footnote}
\usepackage{wrapfig}
\usepackage{pifont}
\usepackage{fontawesome}   

\usepackage{tcolorbox}
\usepackage{tikz}
\usetikzlibrary{tikzmark}
\definecolor{myred}{rgb}{0.7, 0.3, 0.0}
\definecolor{myblue}{HTML}{054488}
\definecolor{mygreen}{HTML}{4CAF50}

\definecolor{BoxFrameColor}{RGB}{100,100,100}
\definecolor{BoxBgColor}{RGB}{248,248,248}
\definecolor{TitleTextColor}{RGB}{255,255,255}
\definecolor{BodyTextColor}{RGB}{50,50,50}
\definecolor{TagColor}{RGB}{41,128,185}        

\definecolor{framecolor}{gray}{0.4}   
\definecolor{bgcolor}{gray}{0.98}     
\definecolor{titlecolor}{gray}{1.0}   
\definecolor{tagcolor}{rgb}{0.2,0.5,0.7} 

\makeatletter
\newcommand*\myfontsize{%
  \@setfontsize\myfontsize{8}{8}%
}
\makeatother

\newcommand{\cmark}{\ding{51}} 
\newcommand{\xmark}{\ding{55}} 

\def\model{AgentBrew\xspace}

\title{AgentBrew: Lifelong Knowledge Brewing from Strong Teachers to Weak LLM Agents}

\author{
  Yangqin Jiang ~~~
  Chao Huang\textsuperscript{}\thanks{Chao Huang is the Corresponding Author.} \\
  The University of Hong Kong \\
  \texttt{\{mrjiangyq99, chaohuang75\}@gmail.com} \\
  \faGithub~\textbf{Github Repo:} \textcolor{blue}{\url{https://github.com/HKUDS/UpSkill}
}
}

\begin{document}

\maketitle
\setcounter{footnote}{0}

\begin{abstract}
Deploying LLM agents typically requires a compact test-time student, even if a stronger teacher is available during training. We study knowledge brewing: distilling a teacher's interactive experience into a persistent external memory for the student. Crucially, this requires no weight updates, expert demonstrations, ground-truth labels, or test-time teacher access. This setting poses two challenges: environments provide only sparse, binary feedback, and teacher-authored notes must be inherently tailored to be concretely executable by a substantially weaker student. To address these hurdles, we propose \textbf{\model}, comprising two coupled components. First, a failure-triggered teacher--Ralph Loop mitigates sparse feedback by transforming student failures into environment-validated notes. Second, student-aware synthesis calibrates teacher knowledge to the weak executor's operational granularity, yielding model-specific, actionable guidance. Extensive evaluations and comprehensive ablations across coding, math, and tool-use tasks demonstrate that this asymmetric, training-free brewing paradigm produces highly capable yet deployable LLM agents.
\end{abstract}


\input{intro}
\input{solution}
\input{evaluation}

\input{relate} 
\input{conclusion}

\clearpage

\bibliographystyle{unsrtnat}
\bibliography{refs}

\input{appendix}

\end{document}

%% file: intro.tex
\section{Introduction}
\label{sec:intro}

Large language model (LLM) agents now solve interactive tasks in coding, math, and tool-use environments~\cite{gou2024tora,trivedi2024appworld, jiang2025lightagent, yang2024swe}. In production, however, the agent that must serve requests is often a small or API-only model: fine-tuning may be unavailable, inference cost must stay low, and model weights are treated as fixed after deployment. A natural alternative is teacher--student agent brewing\footnote{We use \emph{brewing} to denote the offline process of distilling teacher knowledge into a persistent external memory, validated on the target student. By \emph{notes} we mean structured, retrievable corrective rules---conceptually similar to hints or procedural knowledge in prior work, but specifically validated to be executable by the student.}: a stronger teacher agent $\mathcal{T}$ (large LLM) distills interaction experience during a training period, while a compact student agent $\mathcal{S}$ (small LLM) runs alone at test time with only retrieved external knowledge. The central question is how to convert brewing-time interaction into durable notes that raise $\mathcal{S}$'s effective capability without updating its parameters.

\begin{wrapfigure}{r}{0.53\textwidth}
    \centering
    \vspace{-0.15in}
    \includegraphics[width=0.5\textwidth]{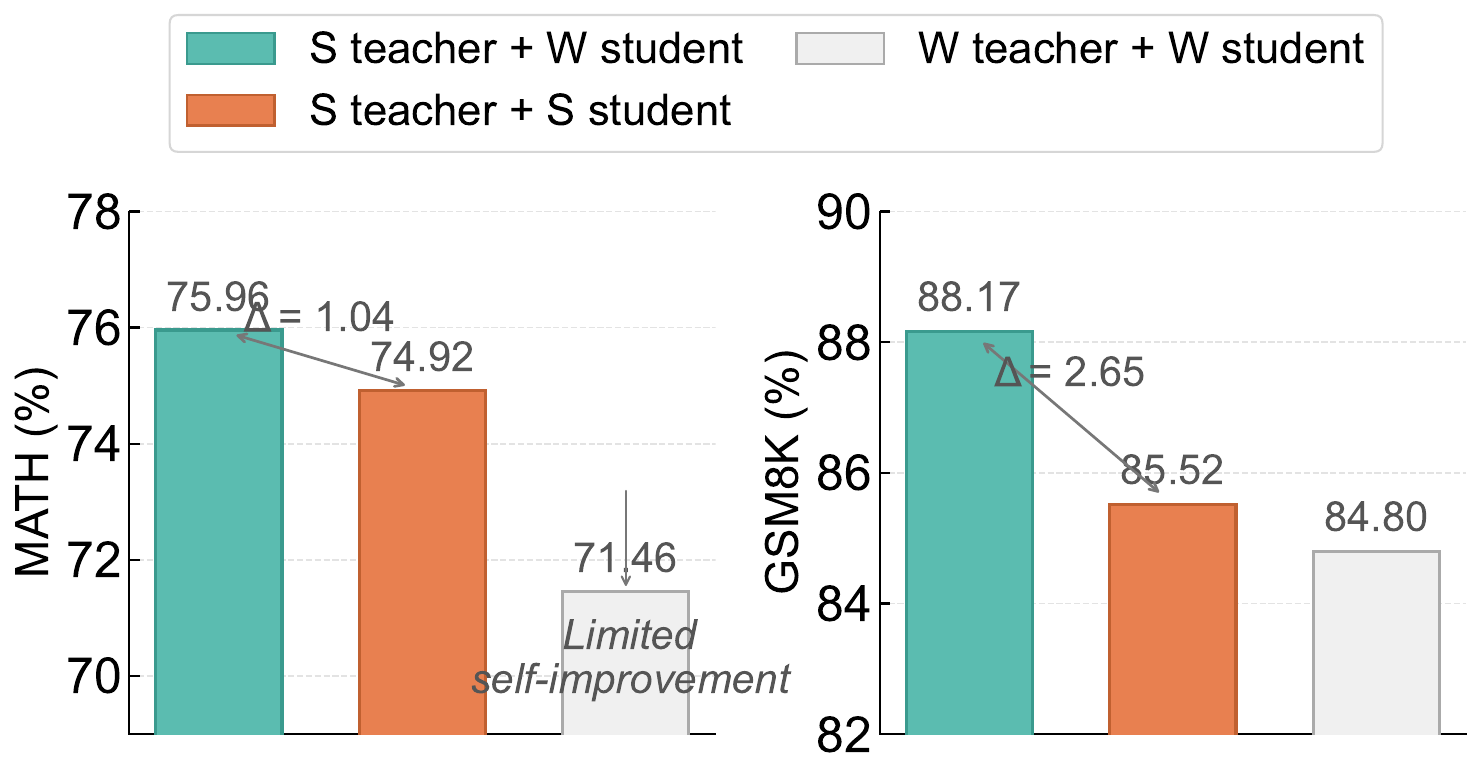}
    \vspace{-0.1in}
    \caption{Accuracy on \textsc{MATH} and \textsc{GSM8K} under different brew-time teacher--student LLM pairings (test student: Qwen3-14B). Strong (S)/weak (W) denote DeepSeek-Chat-v3.1 and Qwen3-14B.}
    \label{fig:math_gsm8k_teacher_student}
    \vspace{-0.1in}
\end{wrapfigure}

Two challenges make this setting substantially harder than either self-evolution or standard distillation alone. \textbf{Challenge~I: sparse environment feedback.} Real benchmarks and deployed APIs typically expose only a terminal binary reward $R \in \{0,1\}$---pass or fail---together with the agent's own rollout, but no ground-truth solution, expert trajectory, or step-level label. When $\mathcal{S}$ fails, the only extra signal is the failure trajectory $\tau_{\mathcal{S}}(x)$; when it succeeds, no teacher intervention is needed. 
\textbf{Challenge~II: teacher--student capability gap (\textit{i.e.}, knowledge is model-specific).} In the deployment scenario above, $\mathcal{T}$ is intentionally much stronger than $\mathcal{S}$, yet the notes stored in memory must be written by $\mathcal{T}$ and executed by $\mathcal{S}$. Knowledge that reads well to a strong model often fails in practice for a weak one: reasoning steps that $\mathcal{T}$ treats as obvious are omitted, vocabulary and concepts exceed $\mathcal{S}$'s comfort zone, and corrective rules are too coarse to translate into concrete actions. Figure~\ref{fig:math_gsm8k_teacher_student} makes this dependency explicit: with Qwen3-14B fixed at test time, \textit{S teacher + W student} brewing outperforms \textit{S teacher + S student} brewing, showing that transferable notes depend on the student model involved in brewing---not merely on teacher quality. Bridging this gap is therefore a first-order design problem.

Existing work addresses only parts of the problem. Self-evolving agents~\cite{shinn2023reflexion,zhao2024expel} can improve from environment feedback without weight updates, but they usually (i) evolve the same model that both acts and learns, (ii) continue adapting at test time, and (iii) are not built for asymmetric teacher-student "brewing", where notes must be validated for a fixed weak executor. Agent distillation~\cite{liu2025structured,qiu2025agentdistill} can compress teacher behavior into trajectories or reusable artifacts, but it typically depends on labeled pairs or successful teacher rollouts, and it transfers knowledge at a teacher-level abstraction rather than as student-executable notes certified to work on the weak executor. 
Accordingly, we organize our investigation around three questions:

\noindent\textbf{Q1 (Learning signal).}
When $\mathrm{Env}$ exposes only binary $R$ and no expert data, how can a weak agent extract useful knowledge from failures, and how can we verify that knowledge before storing it?

\noindent\textbf{Q2 (Capability gap).}
How can a strong teacher author knowledge that a substantially weaker student can execute---not merely parse---given that both generation and validation are biased by the gap?

\noindent\textbf{Q3 (Lifelong accumulation).}
How can validated, student-calibrated knowledge accumulate across tasks in external memory-without updating student weights $\mathcal{S}$-while keeping test-time inference lightweight and stable?

We propose a brew--serve framework (Figure~\ref{fig:framework}) that maps Q1--Q3 to three coupled components: a failure-triggered \textbf{teacher--Ralph Loop}, \textbf{student-aware note synthesis}, and a \textbf{lifelong memory} $\mathcal{M}$. 
We call it \textbf{\model}: \emph{brewing} is the Ralph Loop~\cite{huntley2025ralph} that iteratively distills certified notes from failures, and student-aware synthesis tailors each note to its executor as a brewer adapts method to the ingredients.
Brewing on $\mathcal{D}_{\mathrm{train}}$ proceeds as follows. Student $\mathcal{S}$ rolls out each task in $\mathrm{Env}$; when $R{=}0$, teacher $\mathcal{T}$ reads $(x, \tau_{\mathcal{S}}(x))$---the only supervision beyond binary pass/fail---and drafts a structured candidate note. The note is authored with student-aware prompting and an action-oriented schema so that $\mathcal{T}$ targets $\mathcal{S}$'s reading level and operational granularity rather than its own. The Ralph Loop then stages the note and re-rolls $\mathcal{S}$ on the same task: recovery ($R{=}1$) certifies the note using environment feedback alone, and rejects guidance that remains plausible to $\mathcal{T}$ yet unusable for the weak executor. A curator deduplicates certified notes and appends them to $\mathcal{M}$, which grows across the training stream without updating $\mathcal{S}$. Serving on $\mathcal{D}_{\mathrm{test}}$ reads from frozen $\mathcal{M}^\star$: $\mathcal{T}$ is inactive, $\mathcal{S}$ retrieves skill-scoped top-$k$ notes, and completes each task in a single rollout with no iterative reflection or online memory writes. \model is guided by a simple principle: \textbf{knowledge is not universal---it must fit the agent that executes it}. Our main contributions are:
\begin{itemize}[leftmargin=*]
    \item \textbf{Brew--serve under sparse feedback.}
    We develop a lifelong knowledge-building method for a fixed weak agent using only binary feedback from the environment, without expert trajectories, test-time labels, or any weight updates.
    
    \item \textbf{Student-calibrated transfer.}
    We introduce student-aware note synthesis that proactively authors teacher knowledge at $\mathcal{S}$'s operational granularity, paired with Ralph recovery as a reactive filter that certifies notes on the weak executor rather than on teacher plausibility alone.
    
    \item \textbf{Comprehensive Evaluation.}
    We benchmark \model on three typical agent tasks (\textit{i.e.,} coding, math, and tool-use) under a shared brew--serve protocol, with ablations study assessing the effectiveness of our methods.
\end{itemize}

%% file: solution.tex
\section{The \model Framework}
\label{sec:solution}

\subsection{Problem Formulation}
\label{sec:problem_formulation}

\noindent\textbf{Setup.}
We examine interactive tasks $x$ (\textit{e.g.,} math, coding, or tool-use). A fixed student agent $\mathcal{S}$, powered by a small LLM, interacts with an environment $\mathrm{Env}$ for up to $H$ steps to yield a trajectory $\tau_{\mathcal{S}}(x)$ and a binary terminal reward $R(x,\tau) \in {0,1}$. Notably, $\mathrm{Env}$ provides no ground-truth solutions, labels, or expert trajectories. During training ($\mathcal{D}{\mathrm{train}}$), a stronger teacher agent $\mathcal{T}$, driven by a large LLM, is introduced: upon any failure by $\mathcal{S}$, $\mathcal{T}$ analyzes the task and the failed trajectory to synthesize structured notes into an external memory bank $\mathcal{M}$. During testing ($\mathcal{D}{\mathrm{test}}$), $\mathcal{M}$ is frozen and $\mathcal{T}$ is entirely absent; instead, $\mathcal{S}$ solves tasks in a single pass augmented by notes retrieved from $\mathcal{M}$. Overall, learning is driven purely by environment feedback and failure trajectories, without offline demonstrations, test-time oracle labels, or updates to the parameters of $\mathcal{S}$.

\noindent\textbf{Objective.}
Brew a persistent memory $\mathcal{M}^\star$ on $\mathcal{D}_{\mathrm{train}}$ that maximizes held-out success:
\begin{equation}
\label{eq:objective}
    \mathcal{M}^\star \in \arg\max_{\mathcal{M}} \;
    \mathbb{E}_{x \sim \mathcal{D}_{\mathrm{test}}}\!\left[
        R\bigl(x,\, \tau_{\mathcal{S}}(x \mid \mathcal{M})\bigr)
    \right],
\end{equation}
where $\tau_{\mathcal{S}}(x \mid \mathcal{M})$ is the trajectory rolled out by student agent $\mathcal{S}$ with the top-$k$ notes from $\mathcal{M}$ prepended to its prompt. All lifelong gain is stored in $\mathcal{M}^\star$, not in the student agent's weights.

\subsection{Framework Overview}
\label{sec:framework_overview}

\begin{figure*}[h]
    \centering
    \includegraphics[width=0.95\linewidth]{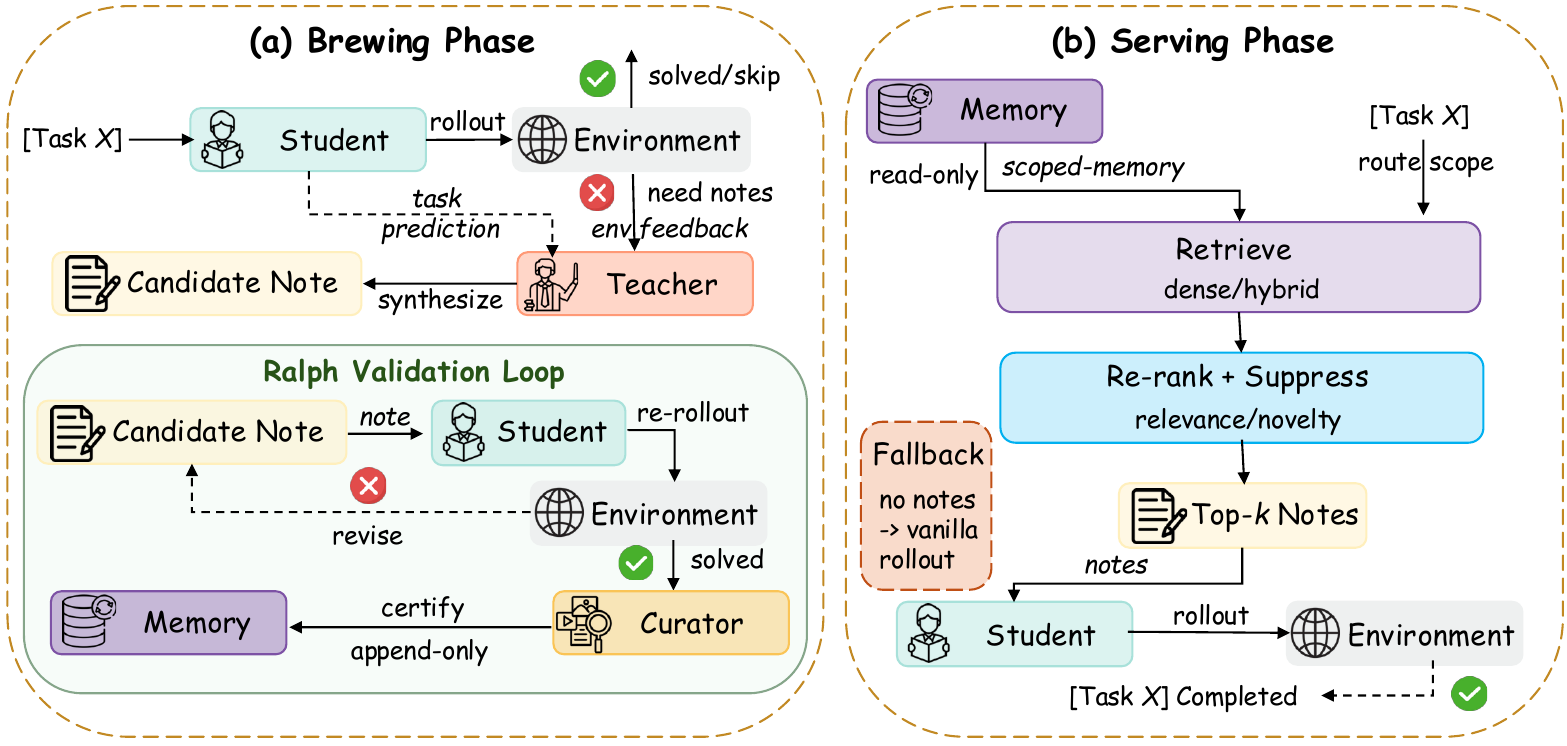}
    \vspace{-0.05in}
    \caption{Overall framework of the proposed \model.}
    \vspace{-0.1in}
    \label{fig:framework}
\end{figure*}

\model operationalizes the objective in Sec.~\ref{sec:problem_formulation} via a \emph{brew--serve} paradigm (Figure~\ref{fig:framework}):
The full system comprises three design choices, each addressing a research question raised in the introduction:
\textbf{(Q1)} a failure-triggered \textbf{teacher} turns sparse binary feedback from $\mathrm{Env}$ into structured candidate notes, with a \textbf{curator} merging environment-certified entries into $\mathcal{M}$ (Sec.~\ref{sec:teacher_ralph});
\textbf{(Q2)} \textbf{reactive Ralph validation} plus \textbf{proactive student-aware synthesis} bridge the capability gap between teacher and student agents (Sec.~\ref{sec:student_aware}); and
\textbf{(Q3)} a \textbf{lifelong memory} $\mathcal{M}$ accumulates validated notes across tasks without updating the student LLM (Sec.~\ref{sec:lifelong_memory}). Here we summarize how they interact:

\noindent $\bullet$ \textbf{Brewing phase} \emph{(on $\mathcal{D}_{\mathrm{train}}$).} For each task $x$, the student agent $\mathcal{S}$ attempts $x$ in $\mathrm{Env}$. If $R{=}1$, the teacher is not invoked and the pipeline advances to the next task. If $R{=}0$, the teacher agent $\mathcal{T}$ reads $(x, \tau_{\mathcal{S}}(x))$ and, via student-aware prompting, proposes a structured candidate note~$n$. The Ralph Loop~\cite{huntley2025ralph} injects $n$ into $\mathcal{S}$ and re-rolls out on the \emph{same} task~$x$: a successful re-rollout ($R{=}1$) validates $n$, whereas persistent failure triggers revision or discard. Notes that pass the Ralph Loop are merged by a curator agent---performing deduplication and quality gating only, as correctness is already certified by the environment---and appended to $\mathcal{M}$. This cycle repeats over the training stream, progressively brewing $\mathcal{M}^\star$.

\noindent $\bullet$ \textbf{Serving phase} \emph{(on $\mathcal{D}_{\mathrm{test}}$).} At test time, memory is read-only and the teacher agent $\mathcal{T}$ is inactive. Given a new task $x$, $\mathcal{S}$ retrieves the top-$k$ notes from frozen memory $\mathcal{M}^\star$ (via skill-scoped routing and reranking), prepends them to its prompt, and completes $x$ in a \emph{single} environment rollout-without an iterative reflection loop or additional validation calls. If retrieval returns no notes, $\mathcal{S}$ falls back to its vanilla policy.

\subsection{From Binary Feedback to Validated Knowledge}
\label{sec:teacher_ralph}

The brewing phase must extract durable knowledge from failures when $\mathrm{Env}$ exposes only a terminal reward $R \in \{0,1\}$: no step-level labels, no expert trajectories, and no oracle explanations. We address \textbf{Q1} by treating each failure as a supervision event. Here we detail the failure-triggered activation pipeline, the note representation written by $\mathcal{T}$, and curation before durable writes.

\subsubsection{Failure-Triggered Teacher Activation}
\label{sec:teacher_activation}

\begin{algorithm}[h]
\caption{Failure-triggered teacher activation on $\mathcal{D}_{\mathrm{train}}$.}
\label{alg:teacher_activation}
\small
\DontPrintSemicolon
\SetFillComment
\KwIn{$\mathcal{D}_{\mathrm{train}}$, student $\mathcal{S}$, teacher $\mathcal{T}$, memory $\mathcal{M}$, environment $\mathrm{Env}$}
\For{each $x \in \mathcal{D}_{\mathrm{train}}$}{
    $\mathcal{N} \gets \textsc{Retrieve}(\mathcal{M}, x)$\tcp*[r]{retrieved notes for $x$ (may be $\emptyset$)}
    $\tau_{\mathcal{S}}(x) \gets \textsc{Rollout}(\mathcal{S}, x, \mathcal{N})$\;
    $R \gets \textsc{Eval}(\mathrm{Env}, x, \tau_{\mathcal{S}}(x))$\tcp*[r]{$R \in \{0,1\}$; sole external supervision from $\mathrm{Env}$}
    \If{$R = 1$}{
        \textbf{continue}\tcp*[r]{success: $\mathcal{T}$ not invoked}
    }
    \Else{
        $n \gets \textsc{DraftNote}(\mathcal{T}, x, \tau_{\mathcal{S}}(x), \textsc{fail})$\tcp*[r]{no reference solution}
        \textsc{RalphLoop}$(x, n, \mathcal{S}, \mathcal{T}, \mathcal{C}, \mathcal{M}, \mathrm{Env})$\tcp*[r]{refinement and curation}
    }
}
\end{algorithm}

During brewing, each task $x \in \mathcal{D}_{\mathrm{train}}$ begins with a rollout by student agent $\mathcal{S}$, producing $\tau_{\mathcal{S}}(x)$ and reward $R \in \{0,1\}$. Binary $R$ is the sole external supervision signal from $\mathrm{Env}$: no step-level labels, no expert trajectories, and no oracle explanations are provided. If $R=1$, the pipeline advances to the next task without invoking $\mathcal{T}$. If $R=0$, we trigger $\mathcal{T}$ on the tuple $(x,\, \tau_{\mathcal{S}}(x),\, \textsc{fail})$. Unlike Supervised Fine-Tuning (SFT), $\mathcal{T}$ never conditions on a reference solution; it diagnoses the failure mode from the student trajectory and task text alone, then proposes a corrective note. This design makes learning possible in settings where only pass/fail feedback is available. Algo.~\ref{alg:teacher_activation} summarizes the control flow.

\subsubsection{Structured Note Generation}
\label{sec:note_generation}

Given a failure, $\mathcal{T}$ outputs a candidate note $n$ as structured JSON rather than a worked example. Each note is an abstract, reusable rule intended for retrieval on future tasks with similar structure:
\begin{itemize}[leftmargin=*]
    \item \texttt{note\_key}: unique identifier (reused when updating an existing rule);
    \item \texttt{trigger\_pattern}: task features that should activate the note;
    \item \texttt{corrective\_rule}: an executable fix directive for $\mathcal{S}$;
    \item \texttt{minimal\_steps}: a short ordered checklist lowering interpretation cost for a weak LLM;
    \item \texttt{tags}: domain or skill labels for routing;
    \item \texttt{quality\_score}: initial confidence, later updated from validation statistics.
\end{itemize}

Notes deliberately avoid storing full solutions to $x$; they compress failures into transferable guardrails. Granularity and wording are calibrated to the student agent's capability.

\subsubsection{Note Curator}
\label{sec:curator}

When the Ralph Loop returns $\mathit{validated}=\textsc{true}$, Algo.~\ref{alg:teacher_activation} passes the environment-certified note to curator agent $\mathcal{C}$ before any durable write. $\mathcal{C}$ does \emph{not} re-judge factual correctness---recovery in $\mathrm{Env}$ already certifies utility to $\mathcal{S}$ on the brewing task $x$. Its role is strictly operational on the \emph{memory store}:

\noindent \textbf{(i) \emph{Deduplication:}} compare $n$ against existing entries in $\mathcal{M}$ via lexical and semantic similarity; merge near-duplicates by updating an existing \texttt{note\_key} instead of appending redundant rules.

\noindent \textbf{(ii) \emph{Quality tracking:}} adjust \texttt{quality\_score} from historical validation outcomes (acceptances, recoveries, and downstream usage).

\noindent \textbf{(iii) \emph{Write gating:}} persist $n$ to $\mathcal{M}$ only if it passes redundancy and quality thresholds; otherwise discard despite a successful Ralph Loop round.

Accepted notes are appended under an append-only policy, extending $\mathcal{M}$ without overwriting prior knowledge or modifying the student LLM.

\subsection{Bridging the Capability Gap}
\label{sec:student_aware}

Teacher agent $\mathcal{T}$ is substantially stronger than student agent $\mathcal{S}$, leading to the capability gap~\cite{li2025small} between to agents. 
We address \textbf{Q2} with a \emph{two-stage} design.
\textbf{First, reactive validation} : the \textbf{Ralph Loop}~\cite{huntley2025ralph} re-rolls $\mathcal{S}$ on the failed task with a candidate note staged in memory and accepts $n$ only when $\mathrm{Env}$ returns $R{=}1$, using terminal reward---not teacher or student self-judgment---as the certification signal.
\textbf{Second, proactive synthesis} : student-aware prompting and note schema calibrate what $\mathcal{T}$ writes before validation, so proposals match $\mathcal{S}$'s reading level and operational granularity.
Ralph provides the environment-grounded filter; proactive synthesis improves validation yield and transferability.

\noindent $\bullet$ \textbf{Why same-level generation fails.}
\label{sec:capability_gap}
Prior work on knowledge transfer, such as ACE~\cite{zhang2025agentic}, assumes that the knowledge producer and consumer have comparable reasoning ability. When $\mathcal{T}$ is substantially stronger than $\mathcal{S}$, this assumption breaks down~\cite{chen2025unveiling, mirzadeh2020improved}: notes often omit intermediate reasoning, use mismatched vocabulary, or specify steps at a granularity too coarse for $\mathcal{S}$ to follow. Such notes may appear plausible to $\mathcal{T}$ yet leave $\mathcal{S}$ unable to recover.

\subsubsection{Reactive Validation: Ralph Loop}
\label{sec:ralph_loop}

Teacher-generated notes can be plausible yet useless or harmful to $\mathcal{S}$. Self-judgment~\cite{shinn2023reflexion} by $\mathcal{T}$ or $\mathcal{S}$ is insufficient because capability gaps bias both generation and critique. 
The Ralph Loop is the environment-validation stage of brewing: it asks whether a candidate note actually helps $\mathcal{S}$ recover on the failed task, using only terminal rewards from $\mathrm{Env}$.

\noindent \textbf{Provisional staging.} After $\mathcal{T}$ proposes $n$, $\textsc{AddProvisional}(\mathcal{M}, n)$ places $n$ in a \emph{provisional} slot---visible to $\textsc{Retrieve}(\mathcal{M}, x)$ during re-attempts on $x$, but not yet part of durable long-term memory. Provisional entries simulate ``what if this note were already in $\mathcal{M}$?'' for the validation rollouts; $\textsc{RollbackProvisional}$ removes the slot if validation fails, and $\textsc{Commit}$ promotes a curated note only after $\mathcal{C}$ approves persistence.

\noindent \textbf{Environment validation.} With $n$ staged provisionally, we re-run $\mathcal{S}$ on the \emph{same} task $x$, retrieving $\mathcal{N}$ that includes $n$ along with any notes already committed for $x$.
Let $\tau_{\mathcal{S}}^{(r)}(x \mid n)$ denote the $r$-th re-attempt trajectory and $R^{(r)}$ its terminal reward.
The Ralph Loop accepts $n$ when recovery occurs:
\begin{equation}
\small
\label{eq:ralph_accept}
    R^{(r)} = 1 \quad \text{for some re-attempt } r \in \{1,\ldots,R_{\max}\},
\end{equation}
where $R_{\max}$ caps extra environment calls.
If no attempt succeeds, $\textsc{RollbackProvisional}(\mathcal{M})$ discards the staging slot; optionally $\mathcal{T}$ revises $n$ from the latest failed trajectory $\tau^{(r)}$ before the next $r$.
On success, the loop returns $\mathit{validated}=\textsc{true}$ and leaves $n$ provisionally staged for $\mathcal{C}$; Algo.~\ref{alg:teacher_activation} then invokes curation.

\begin{algorithm}[h]
\caption{Ralph Loop: environment validation of a candidate note on task $x$.}
\label{alg:ralph_loop}
\small
\SetAlgoLined
\DontPrintSemicolon
\SetSideCommentLeft
\SetFillComment
\KwIn{task $x$, candidate note $n$, agents $\mathcal{S}$, $\mathcal{T}$, memory $\mathcal{M}$, environment $\mathrm{Env}$, budget $R_{\max}$}
\KwOut{$\mathit{validated} \in \{\textsc{true}, \textsc{false}\}$}
$\textsc{AddProvisional}(\mathcal{M}, n)$\tcp*[t]{stage $n$ for $\textsc{Retrieve}$ only; not durable yet}
$\mathit{validated} \gets \textsc{false}$\;
\For{$r = 1$ \KwTo $R_{\max}$}{
    $\mathcal{N} \gets \textsc{Retrieve}(\mathcal{M}, x)$\;
    $\tau^{(r)} \gets \textsc{Rollout}(\mathcal{S}, x, \mathcal{N})$\;
    $R^{(r)} \gets \textsc{Eval}(\mathrm{Env}, x, \tau^{(r)})$\;
    \If{$R^{(r)} = 1$}{
        $\mathit{validated} \gets \textsc{true}$\;
        \textbf{break}\tcp*[t]{satisfies Eq.~\ref{eq:ralph_accept}}
    }
    \If{$r < R_{\max}$}{
        $n \gets \textsc{DraftNote}(\mathcal{T}, x, \tau^{(r)}, \textsc{fail})$\tcp*[t]{revise $n$ from failed re-attempt}
        $\textsc{UpdateProvisional}(\mathcal{M}, n)$\tcp*[t]{replace staged copy only}
    }
}
\If{$\neg\mathit{validated}$}{
    $\textsc{RollbackProvisional}(\mathcal{M})$\tcp*[t]{no recovery; discard staged $n$}
}
\Return{$\mathit{validated}$}\tcp*[t]{on success, $n$ remains provisional for $\mathcal{C}$}
\end{algorithm}

Algo.~\ref{alg:ralph_loop} details the validation procedure.
It uses only $R^{(r)}$ from $\mathrm{Env}$---never ground-truth labels on $\mathcal{D}_{\mathrm{test}}$ and never oracle answers during serving.
At test time, no Ralph Loop calls are made; validated knowledge is consumed from the frozen $\mathcal{M}^\star$ in a single rollout.

\subsubsection{Proactive Synthesis:Student-Aware Teacher Prompting}
\label{sec:proactive_synthesis}

Ralph validation alone is a costly filter: when $\mathcal{T}$ drafts ACE-style bullets that omit steps $\mathcal{S}$ cannot infer, most proposals fail within $R_{\max}$. We therefore instruct $\mathcal{T}$ to treat $\mathcal{S}$ as the sole intended reader: notes must read like a checklist for a junior agent, not an expert post-mortem. Given $(x,\, \tau_{\mathcal{S}}(x))$ from a failed rollout, $\mathcal{T}$ follows a three-step synthesis protocol before emitting JSON:

\noindent \textbf{(i) \emph{Failure-mode identification:}} Parse $\tau_{\mathcal{S}}(x)$ and the environment report to label the dominant error type (\textit{e.g.}, wrong API call order, sign error, misread constraint).

\noindent \textbf{(ii) \emph{Capability-gap localization:}} Attribute the error to a \emph{specific} missing skill or knowledge fragment at $\mathcal{S}$'s level---not a vague ``logical mistake''---distinguishing representation errors from execution errors where possible.

\noindent \textbf{(iii) \emph{Executable rule extraction:}} Emit a corrective procedure in imperative form (``first do $A$, then $B$, watch for $C$''), banning appeals to tacit understanding (``deeply analyze $X$'') that $\mathcal{S}$ cannot operationalize.

The prompt also states $\mathcal{S}$'s model class and known weaknesses (\textit{e.g.}, limited multi-step arithmetic, brittle tool syntax), biasing $\mathcal{T}$ toward shorter sentences, explicit ordering, and surface-level triggers grounded in $\tau_{\mathcal{S}}(x)$. 


\subsection{Lifelong Memory}
\label{sec:lifelong_memory}

\model addresses \textbf{Q3} by treating external memory $\mathcal{M}$ as the sole durable state: validated notes accumulate across $\mathcal{D}_{\mathrm{train}}$, and student agent $\mathcal{S}$ reads from a frozen $\mathcal{M}^\star$ at test time. No gradient updates or adapter weights are applied; effective capability grows because $\mathcal{M}$ grows. We structure $\mathcal{M}$ to limit cross-domain interference, append new knowledge without erasing old entries, and retrieve notes with a serving pipeline tuned for single-pass inference.

\subsubsection{Skill-Scoped Memory Architecture}
\label{sec:memory_architecture}

A monolithic note pool induces \emph{cross-scope interference}: notes from one skill are retrieved on unrelated tasks. We therefore partition $\mathcal{M}$ into disjoint skill scopes $\{\mathcal{M}_s\}_{s \in \mathcal{S}}$, each indexing notes for a single capability or environment slice. Every note stores \texttt{tags} and an optional scope identifier; during curation, the teacher writes it to the corresponding $\mathcal{M}_s$. At test time, retrieval first selects a scope subset $\mathcal{M}_{\mathrm{scope}} \subseteq \mathcal{M}$ and ranks notes only within that subset-never over the full pool.

Scope routing is orthogonal to note ranking. When $x$ does not clearly specify a scope, we choose $\mathcal{M}_{\mathrm{scope}}$ using three signals, in order: \textbf{(i)} explicit scope or tool names in $x$; \textbf{(ii)} lexical or semantic similarity between $x$ and a scope-description catalog maintained alongside $\mathcal{M}$; \textbf{(iii)} actions observed in the student rollout $\tau_{\mathcal{S}}(x)$ when (i)–(ii) are inconclusive.

\subsubsection{Continual Accumulation and Stability}
\label{sec:continual_accumulation}

During brewing, accepted notes are merged with \textbf{append-only} semantics: curator deduplication may \emph{update} an existing \texttt{note\_key}, but validated rules are never deleted to make room for new ones, avoiding catastrophic forgetting of earlier skills.
Each note record tracks usage statistics and \texttt{quality\_score} so frequently harmful or redundant entries can be down-weighted at retrieval time without erasing history.

Train and test streams are strictly separated. While processing $\mathcal{D}_{\mathrm{train}}$, $\mathcal{M}$ is writable. On $\mathcal{D}_{\mathrm{test}}$, $\mathcal{M}^\star$ is \textbf{frozen read-only}: no teacher calls, no provisional writes, and no statistic updates that leak test-task content back into the store. This guarantees that performance on held-out tasks measures generalization from brewed knowledge, not online adaptation to test labels or test-task feedback.

\subsubsection{Test-Time Knowledge-Augmented Inference}
\label{sec:test_time_inference}

Serving reuses the objective in Eq.~\ref{eq:objective} with a fixed retrieval-and-rollout procedure (Appendix~\ref{app:test_time}). Given task $x$, we encode a query $q(x)$ from the instruction, select the skill scope, and retrieve candidate notes from $\mathcal{M}^\star_{\mathrm{scope}}$ using dense or hybrid lexical--semantic scoring. Candidates are reranked by
\begin{equation}
\label{eq:note-rank}
    \mathrm{score}(n, x) = \mathrm{sim}(q(x), n) \cdot \rho(n) \cdot \phi(n),
\end{equation}
where $\mathrm{sim}(\cdot)$ is retrieval similarity, $\rho(n)$ is the stored \texttt{quality\_score} (and success-rate prior), and $\phi(n)$ is a frequency penalty that down-weights notes over-selected in past tasks to promote diversity. Rather than hard-dropping low-scoring notes, \textbf{soft suppression} applies $\phi(n)$ and a relative-score floor: notes below a fraction of the top score are dampened but may still appear when the pool is sparse, reducing brittle all-or-nothing retrieval. The top-$k$ notes are formatted and prepended to $\mathcal{S}$'s prompt; $\mathcal{S}$ then performs one $\textsc{Rollout}$ in $\mathrm{Env}$. If retrieval returns an empty set, $\mathcal{S}$ falls back to its vanilla policy with no memory augmentation.


%% file: evaluation.tex
\section{Evaluation}
\label{sec:eval}

\subsection{Experimental Setup}

\subsubsection{Tasks and Benchmarks}
We evaluate \model across three representative agent scenarios---\textbf{coding}, \textbf{math}, and \textbf{tool-use}---each probing a distinct capability profile: synthesizing executable programs from natural-language specifications, performing multi-step symbolic reasoning, and completing long-horizon tasks through interactive API calls. More details are in App.~\ref{app:benchmark}.

\noindent $\bullet$ For the \textbf{coding} tasks, we evaluate \model on \textit{MBPP}~\cite{austin2021program}, a benchmark of roughly 1,000 entry-level Python programming tasks, and \textit{MBPP+}~\cite{liu2023your}, an extension that applies stricter unit testing to a curated subset to reduce spurious passes. 

\noindent $\bullet$ For the \textbf{math} tasks, we evaluate \model on \textit{MATH}~\cite{hendrycks2021measuring}, comprising challenging competition-level problems, and \textit{GSM8K}~\cite{cobbe2021training}, which tests multi-step grade-school arithmetic reasoning. We score final-answer accuracy using the Qwen2.5-Math~\cite{yang2024qwen2} evaluation framework.

\noindent $\bullet$ For the \textbf{tool-use} tasks, we evaluate \model on \textit{AppWorld}~\cite{trivedi2024appworld}, an interactive benchmark where agents write and execute Python code to complete realistic, multi-app daily tasks in a simulated environment with multiple APIs.

\subsubsection{Models, Baselines and Metrics}
\noindent $\bullet$ \textbf{LLM Models.}
In our experiments, we use \texttt{deepseek-chat-v3.1}~\cite{liu2024deepseek} by default for the teacher agent and \texttt{qwen3-14b}~\cite{yang2025qwen3} by default for the student agent.

\noindent $\bullet$ \textbf{Baseline Methods.}
The main baseline methods we compare against include: \textbf{Simple RAG}, \textbf{ExpeL}~\cite{zhao2024expel}, \textbf{Reflexion}~\cite{shinn2023reflexion}, \textbf{ACE}~\cite{zhang2025agentic}, \textbf{ReAct}~\cite{yao2023react}, and \textbf{LoRA}~\cite{hu2022lora}.
Among these, \textbf{Simple RAG} refers to saving the successful trajectories of the teacher agent during training, and directly retrieving these trajectories during testing to help the student agent complete the task.

\noindent $\bullet$ \textbf{Evaluation Metrics.}
We use accuracy for math tasks, the strict EvalPlus~\cite{liu2023your} benchmark for coding tasks, and Task Goal Completion (TGC) for tool-use tasks (\textit{i.e.}, AppWorld) as our evaluation metrics.

\subsection{Main Performance Evaluation}

\begin{table*}[h]
\centering
\vspace{-0.1in}
\caption{Main Results on Three Agent Scenarios. ($^{*}$ indicates \model significantly outperforms the strongest baseline in Group~I and Group~III at $p{<}0.05$ under a two-proportion $z$-test.)}
\label{tab:baseline_main_results}
\small
\setlength{\tabcolsep}{6pt}
\resizebox{\linewidth}{!}{
\begin{tabular}{lcc|cc|cc|cc}
\toprule
    \multirow{2}{*}{\textbf{Method}} & \multirow{2}{*}{\textbf{Teacher}} & \multirow{2}{*}{\textbf{Test-time}} & \multicolumn{2}{c}{\textbf{Math}} & \multicolumn{2}{c}{\textbf{Coding}} & \multicolumn{2}{c}{\textbf{Tool-Use}} \\
    \cmidrule(lr){4-5}\cmidrule(lr){6-7}\cmidrule(lr){8-9}
    & & & MATH ($\%$) & GSM8K ($\%$) & MBPP ($\%$) & MBPP$+$ ($\%$) & Normal ($\%$) & Challenge ($\%$) \\
    \midrule
    \multicolumn{9}{c}{\textbf{Group I: \textit{Student-Only, Single Rollout}}} \\
    \cmidrule(lr){1-9}
    ReAct & \xmark & 1 rollout & 71.66 & 86.12 & 51.02 & 68.83 & 22.00 & 11.30 \\
    SFT(LoRA)\footnotemark & \xmark & 1 rollout & 61.50 & 87.11 & -- & -- & -- & -- \\
    \midrule
    \multicolumn{9}{c}{\textbf{Group II: \textit{Student-Only, Multi-Rollout + Test-Time Feedback}}} \\
    \cmidrule(lr){1-9}
    Reflexion ($k{=}2$) & \xmark & 2 rollouts + fb & 77.12 & 91.51 & 54.22 & 70.78 & 29.76 & 14.63 \\
    Reflexion ($k{=}3$) & \xmark & 3 rollouts + fb & 80.06 & 93.10 & 60.54 & 70.13 & 36.31 & 17.75 \\
    \midrule
    \multicolumn{9}{c}{\textbf{Group III: \textit{Teacher-Augmented Brewing, Single Rollout}}} \\
    \cmidrule(lr){1-9}
    Simple RAG & \cmark & 1 rollout & 74.32 & 82.56 & 60.54 & 71.43 & 29.17 & 13.19 \\
    ExpeL & \cmark & 1 rollout & 70.60 & 88.10 & 61.22 & 68.18 & 28.57 & 10.80 \\
    ACE & \cmark & 1 rollout & 73.82 & 87.87 & 58.61 & 64.29 & 26.79 & 10.55 \\
    \textbf{\model} & \cmark & 1 rollout & \textbf{75.96}$^{*}$ & \textbf{88.17}$^{*}$ & \textbf{61.22} & \textbf{71.43}$^{*}$ & \textbf{31.55}$^{*}$ & \textbf{15.35}$^{*}$ \\
\bottomrule
\end{tabular}
}
\vspace{-0.1in}
\end{table*}

Table~\ref{tab:baseline_main_results} reports main results on all six benchmarks. Unless noted otherwise, every method uses Qwen3-14B as the test-time \emph{generator}.

\noindent \textbf{(i) Strong-teacher brewing \emph{vs.}\ student-only reflection.}
ReAct and Reflexion (Table~\ref{tab:baseline_main_results}, Groups~I--II) rely entirely on the weak student at test time: ReAct acts in a single pass, while Reflexion adds multiple reflection rounds that re-attempt each task after observing explicit binary feedback from the environment. The Reflexion setup with $k{>}1$ is therefore a stronger inference protocol than our single-rollout serving (Group~III), as it consumes additional test-time trials and outcome signals that are unavailable to all other methods. Even so, \model clearly outperforms ReAct across all three scenarios, and despite using only one test-time attempt, matches or exceeds Reflexion ($k{=}2$) on coding and tool-use---MBPP (61.22\% vs.\ 54.22\%), MBPP+ (71.43\% vs.\ 70.78\%), AppWorld Normal (31.55\% vs.\ 29.76\%), and AppWorld Challenge (15.35\% vs.\ 14.63\%). Reflexion's advantage is concentrated on math benchmarks, where iterative self-correction from explicit pass/fail signals is most beneficial for multi-step arithmetic reasoning. These trends support that routing binary failure feedback through a strong teacher and validating notes before storage yields more useful knowledge than unaided student self-reflection, particularly when weak-model reflection is insufficient for complex code generation and API interaction.

\noindent \textbf{(ii) Student-calibrated transfer \emph{vs.}\ teacher-agnostic memory.}
Simple RAG, ExpeL, and ACE also use a strong teacher to build external memory (Group~III), but none explicitly calibrates that memory to the weak student or validates it on the executor that must use it at test time. \model consistently outperforms ACE across math, coding, and tool-use, and achieves the best or tied-best results against Simple RAG and ExpeL on most benchmarks---with particularly clear gains on harder coding and tool-use splits such as MBPP+ and AppWorld \texttt{test\_challenge}. These results support that bridging the teacher--student capability gap through student-aware authoring and environment-side validation is key to turning teacher knowledge into guidance that a fixed weak agent can actually follow.

\footnotetext{SFT(LoRA) is only evaluated on math, where ground-truth answers are available for supervised training. Coding and tool-use tasks provide only binary pass/fail signals without reference solutions, making supervised fine-tuning infeasible.}

\subsection{Ablation Study}

\begin{wraptable}{r}{0.6\textwidth}
\centering
\vspace{-0.25in}
\caption{Ablation Study on \textsc{Math} Scenario.}
\label{tab:ablation_math}
\small
\setlength{\tabcolsep}{4pt}
\begin{tabular}{lccr}
\toprule
    \textbf{Variant} & MATH ($\%$) & GSM8K ($\%$) \\
    \midrule
    \textit{w/o} Ralph Loop & 75.58 & 84.61 \\
    \textit{w/o} Memory & 71.66 & 86.12 \\
    \textit{w/o} Test-Time Suppression & 74.98 & 84.84 \\
    \textit{w} Strong Student Knowledge & 74.92 & 85.52 \\
    \midrule
    \textbf{\model} & 75.96 & 88.17 \\
\bottomrule
\end{tabular}
\vspace{-0.2in}
\end{wraptable}

To isolate the contribution of individual components, we ablate \model on the math scenario (MATH and GSM8K) while holding the test-time student fixed (\textit{i.e.}, Qwen3-14B). Table~\ref{tab:ablation_math} compares the full system against four variants, each modifying one part of \model.

\noindent\textbf{Variant settings.}
$\bullet$ \textit{w/o Ralph Loop}: The teacher drafts notes upon student failure without performing Ralph recovery checks.
$\bullet$ \textit{w/o Memory}: The external memory is disabled; no notes are stored or retrieved at test time.
$\bullet$ \textit{w/o Test-Time Suppression}: Serving relies solely on brewing-time quality scores, omitting soft suppression during retrieval.
$\bullet$ \textit{w/ Strong Student Knowledge}: A strong LLM (DeepSeek-Chat-V3.1) replaces the weak student during the brewing phase, while test-time inference continues to use Qwen3-14B.

\noindent\textbf{Findings.}
$\bullet$ \textit{w/o Ralph Loop} (MATH: 75.58, GSM8K: 84.61) lags behind the full system, particularly on GSM8K, confirming that recovery checks successfully filter out notes that are unhelpful to the weak student.
$\bullet$ \textit{w/o Memory} yields the lowest MATH score (71.66) while maintaining GSM8K performance (86.12), highlighting that external memory is vital for complex reasoning tasks even if the student excels at basic arithmetic.
$\bullet$ \textit{w/o Test-Time Suppression} (MATH: 74.98, GSM8K: 84.84) underperforms the full system, showing that soft suppression successfully down-weights notes that score highly during training but lack retrieval-time relevance.
$\bullet$ \textit{w/ Strong Student Knowledge} (MATH: 74.92, GSM8K: 85.52) also trails the full system, demonstrating that notes tailored to a strong surrogate student transfer ineffectively to Qwen3-14B during inference.

\subsection{Deep Analysis of Ralph Loop}
\vspace{-0.25in}
\begin{table*}[h]
\centering
\caption{Training-phase statistics for Full Ralph vs.\ \textit{w/o Ralph} on MATH and GSM8K.}
\label{tab:rlaph_analysis}
\small
\setlength{\tabcolsep}{6pt}
\begin{tabular}{lcccc}
\toprule
& \multicolumn{2}{c}{\textbf{MATH}} & \multicolumn{2}{c}{\textbf{GSM8K}} \\
\cmidrule(lr){2-3} \cmidrule(lr){4-5}
\textbf{Metric} & \textbf{Full Ralph} & \textbf{\textit{w/o} Ralph} & \textbf{Full Ralph} & \textbf{\textit{w/o} Ralph} \\
\midrule
Training success rate          & 92.2\%  & 89.6\%  & 91.3\%  & 87.6\%  \\
Failed tasks                   & 78      & 104     & 87      & 124     \\
\midrule
Ralph committed tasks          & 28      & 0       & 44      & 0       \\
Direct-write tasks             & 20      & 28      & 22      & 48      \\
Total write tasks              & 48      & 28      & 66      & 48      \\
\midrule
Note write rate                & 61.5\%  & 26.9\%  & 75.9\%  & 38.7\%  \\
Curator rejections             & 30      & 76      & 21      & 76      \\
Curator rejection rate         & 38.5\%  & 73.1\%  & 24.1\%  & 61.3\%  \\
\midrule
Notes in knowledge base        & 39      & 27      & 43      & 33      \\
\bottomrule
\end{tabular}
\vspace{-0.1in}
\end{table*}

During training, the Ralph Loop directly improves task-solving performance, with Full Ralph achieving success rates of 92.2\% on MATH and 91.3\% on GSM8K (compared to 89.6\% and 87.6\% \textit{w/o Ralph}). More importantly, Ralph fundamentally changes knowledge accumulation. Despite encountering \textit{more} training failures (104 vs.\ 78 on MATH; 124 vs.\ 87 on GSM8K), the \textit{w/o Ralph} variant produces \textit{fewer} final notes (27 vs.\ 39 on MATH; 33 vs.\ 43 on GSM8K). The root cause is the note conversion rate: Full Ralph translates 61.5\% and 75.9\% of failures into committed notes, while \textit{w/o Ralph} converts a mere 26.9\% and 38.7\%.

This efficiency gap arises from differing validation mechanisms. Without Ralph, notes are generated post-hoc without proof of utility, forcing the curator to act as the sole quality gate and reject 61--73\% of them. In Full Ralph, the loop acts as an implicit pre-filter: provisional notes reach the Curator only after demonstrably helping the student recover. This student-validated signal cuts the rejection rate to 24--39\% and guarantees that every committed note is empirically effective. The result is a richer, reliably grounded knowledge base that drives Full Ralph's superior test-time performance.

\subsection{Qualitative Example: Why Brewed Notes Generalize}

A natural concern is whether notes validated by recovery on a specific brew task encode task-specific patches rather than transferable knowledge. The structured note schema is explicitly designed to prevent this: fields such as \texttt{trigger\_pattern}, \texttt{corrective\_rule}, and \texttt{not\_applicable\_when} constrain the teacher to describe \emph{when} and \emph{how} to apply a rule, not to memorize answers. Below is a representative note brewed from a MATH training problem involving factorial simplification.

This note encodes a general algebraic strategy without storing any final numeric answer---the brew task's correct result ($\sqrt{576}=24$) is absent. Instead, it captures the abstract corrective rule (simplify before computing), its precondition (shared factors exist), and the failure signature that should trigger retrieval (premature full computation). The \texttt{not\_applicable\_when} field further constrains the note's scope, preventing retrieval on tasks where the rule does not apply. Quantitatively, this single note was retrieved and used on \textbf{1,200 distinct test tasks} with a \textbf{55.6\% success rate} when applied, confirming that the structured schema successfully distills reusable heuristics rather than task-specific patches.

\begin{tcolorbox}[
  colframe=framecolor,
  colback=bgcolor,
  coltitle=titlecolor,
  colbacktitle=framecolor,
  fonttitle=\bfseries,
  title={Example Brewed Note (MATH, factorial simplification)},
  boxrule=1pt,
  arc=2mm,
  left=8pt,
  right=8pt,
  top=6pt,
  bottom=6pt
]
\small
\renewcommand{\arraystretch}{1.15}
\begin{tabular}{@{}p{0.96\textwidth}@{}}
{\color{tagcolor}\bfseries \small[Trigger]}~~\textit{When should this note be retrieved?}\\[2pt]
The student computes the full factorial before simplifying the expression. \\[3pt]
{\color{tagcolor}\bfseries \small[Corrective rule]}~~\textit{What should the student do instead?}\\[2pt]
Simplify the expression algebraically before performing any calculations to reduce computational complexity and avoid errors. \\[3pt]
{\color{tagcolor}\bfseries \small[Minimal pattern]}~~\textit{How to execute, step by step.}\\[2pt]
Cancel common factors in the numerator and denominator before calculating the full value of the factorial. \\[3pt]
{\color{tagcolor}\bfseries \small[Anti-pattern]}~~\textit{What behavior to avoid.}\\[2pt]
Directly computing the full factorial without simplifying the expression first, which increases the risk of computational errors and unnecessary complexity. \\[3pt]
{\color{tagcolor}\bfseries \small[Not applicable when]}~~\textit{When to suppress this note.}\\[2pt]
When the denominator does not share any common factors with the numerator or when simplification is not possible. \\[3pt]
{\color{tagcolor}\bfseries \small[Evidence]}~~\textit{The failure trajectory that triggered this note.}\\[2pt]
\textit{The student computed $8! = 40320$, divided it by 70 to get 576, and then took the square root to arrive at 24, but incorrectly claimed the result was 64 and 8, suggesting a misunderstanding of the computation steps.} \\
\end{tabular}
\end{tcolorbox}

\subsection{Effectiveness of Knowledge Brewing on Different LLMs}

\begin{wrapfigure}{r}{0.5\textwidth}
\centering
\vspace{-0.1in}
\includegraphics[width=0.99\linewidth]{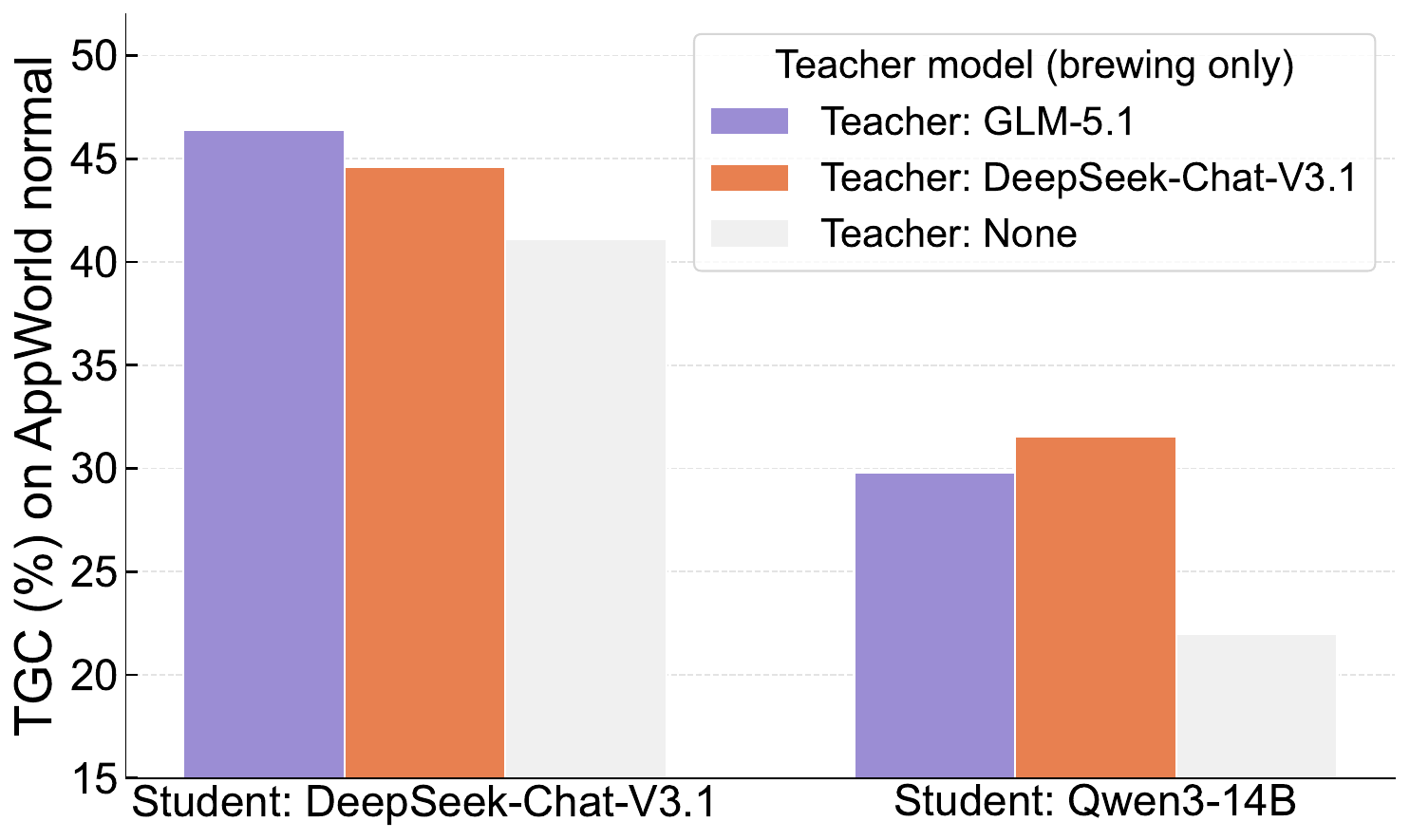}
\caption{Performance on AppWorld \texttt{test\_normal} under different teacher--student LLM pairings.}
\label{fig:llm_generalization}
\vspace{-0.15in}
\end{wrapfigure}

To determine whether \model relies on specific model pairings, we evaluate various combinations of \emph{teachers} (active only on $\mathcal{D}_{\mathrm{train}}$) and \emph{students} (active at brewing and serving) on AppWorld \texttt{test\_normal}, reporting TGC. Holding the rest of the pipeline fixed, we compare three teachers (GLM-5.1~\cite{zeng2026glm}, DeepSeek-Chat-v3.1, and \textit{None}) against two student backbones (DeepSeek-Chat-v3.1 and Qwen3-14B). Figure~\ref{fig:llm_generalization} shows the results, yielding the following insights:

\noindent \textbf{(i) A strong teacher aids diverse LLM pairings.}
Across both student backbones, invoking a strong teacher during brewing strictly outperforms the \textit{None} baseline (where students learn only from their own failures). TGC increases from 41.1\% to 46.4\% for DeepSeek-Chat-v3.1, and from 22.0\% to 31.55\% for Qwen3-14B. These gains confirm that \model robustly generalizes across different vendor and capability combinations.

\noindent \textbf{(ii) Teacher strength does not strictly dictate transfer success.}
While strong teachers consistently beat the baseline, optimal teacher--student pairings vary: DeepSeek-Chat-v3.1 pairs best with Qwen3-14B (31.55\% vs.\ 29.8\%), while GLM-5.1 pairs best with DeepSeek-Chat-v3.1 (46.4\% vs.\ 44.6\%). The smaller gap between teachers on Qwen3-14B likely stems from the student's limited execution capacity---notes that pass brewing may still fail at serving time, making differences in teacher quality harder to realize. This observation strongly motivates \model's student-side validation: rather than assuming teacher-optimal text is universally beneficial, notes must be explicitly validated against the target student's capabilities.

\subsection{Scaling with Training Data}
\label{sec:scaling}

\begin{wrapfigure}{r}{0.5\textwidth}
\centering
\vspace{-0.1in}
\includegraphics[width=0.99\linewidth]{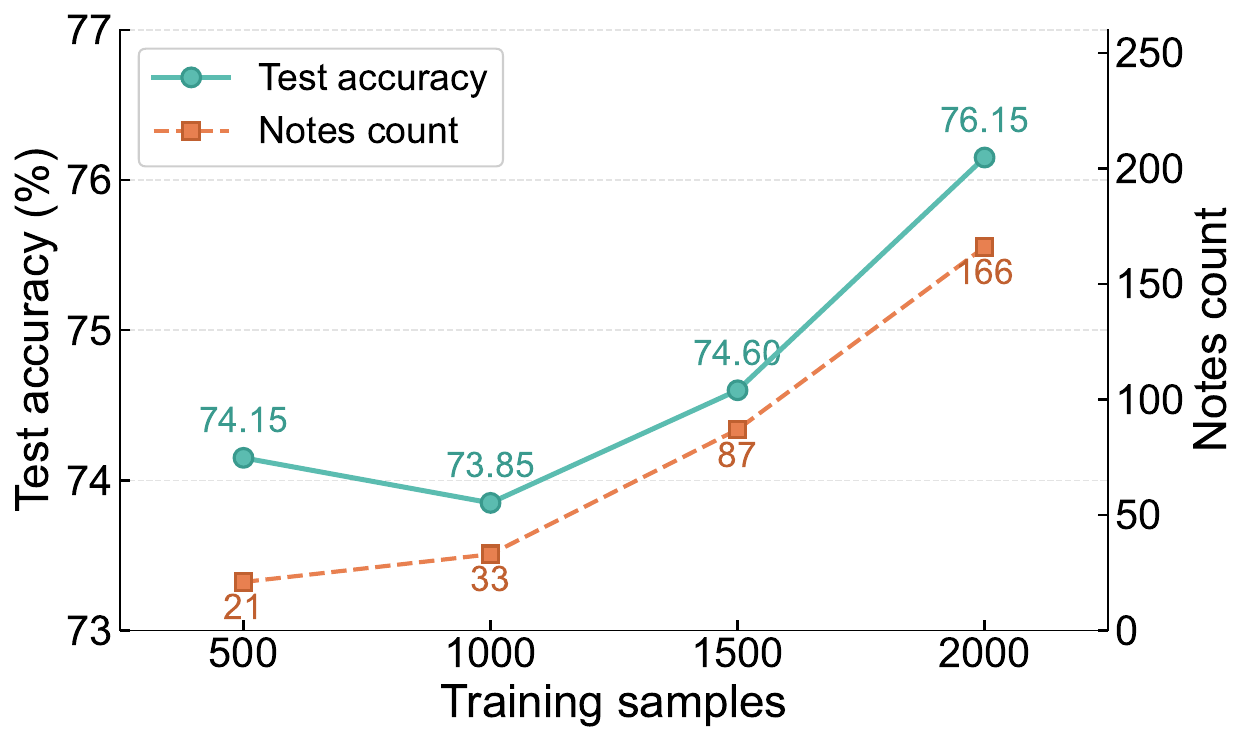}
\caption{Scaling on MATH Benchmark.}
\vspace{-0.15in}
\label{fig:math_scaling}
\end{wrapfigure}

We further ask whether \model exhibits \emph{scaling}: as more training trajectories are brewed, does test-time performance improve? We incrementally brew on subsets of MATH $\mathcal{D}_{\mathrm{train}}$ containing $500$, $1000$, $1500$, and $2000$ samples, respectively, while keeping all other pipeline settings fixed. For each checkpoint, we freeze the resulting memory $\mathcal{M}^\star$ and evaluate it with Qwen3-14B in a single test-time rollout. To control evaluation cost, we report accuracy on a fixed subset of $2000$ sampled MATH test problems rather than the full \texttt{test} split.

Figure~\ref{fig:math_scaling} summarizes the scaling trend. The number of brewed notes grows from 21 to 166 as training exposure increases, indicating that \model continues to distill and retain useful knowledge rather than saturating early. Test accuracy rises from 74.15\% at 500 training problems to 76.15\% at 2000, with a modest dip at 1000 problems (73.85\%) before recovering at larger scales. Overall, the upward trend at 1500--2000 problems suggests that \model benefits from additional brewing data on MATH.

\subsection{Real-World Deployment: A Case Study on Terminal-Bench}
\label{sec:tb_deployment}

To complement the controlled academic benchmarks above, we deploy \model on \textbf{Terminal-Bench 2.0}~\cite{merrill2026terminal}, a benchmark comprising 89 real-world terminal tasks spanning 16 categories including software engineering, system administration, data science, security, and scientific computing. We use a stratified 25/64 train/test split and integrate \model into the Claude Code~\cite{anthropic2025claudecode} agent harness, with \texttt{deepseek-v4-pro}~\cite{xu2026deepseek} as the teacher and \texttt{deepseek-v4-flash} as the student. The full pipeline mirrors our main experiments: student baseline, teacher baseline, per-task brewing, category-level curation, Ralph validation, and final serve.

\begin{figure}[h]
\centering
\includegraphics[width=0.85\linewidth]{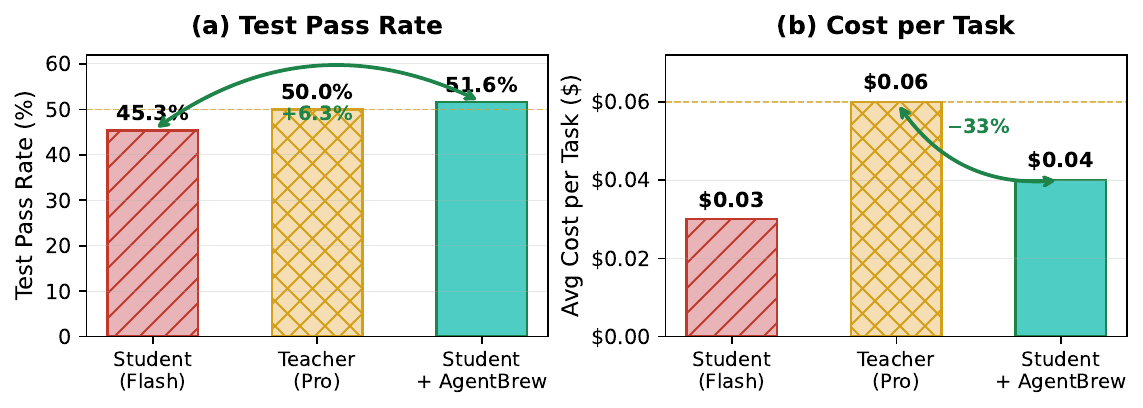}
\caption{Test pass rate and cost per task on Terminal-Bench 2.0 (64 test tasks). (a) \textbf{Left:} pass rate comparison. \model surpasses the teacher model (51.6\% vs.\ 50.0\%) while starting from a substantially weaker student (45.3\%). (b) \textbf{Right:} average cost per task. \model costs \$0.04/task---33\% less than the teacher's \$0.06/task. Dashed lines mark the teacher's reference level.}
\label{fig:tb_main}
\end{figure}

\noindent \textbf{Main Results.}
Figure~\ref{fig:tb_main} summarizes the key findings. The standalone student achieves only 45.3\% pass rate, while the teacher reaches 50.0\%. With \model-brewed skills, the student climbs to \textbf{51.6\%}, surpassing the stronger teacher while costing \$0.04 per task---a 33\% reduction from the teacher's \$0.06. The one-time brewing cost across all training tasks totals approximately \$116, which is amortized over unlimited downstream deployments.

\noindent \textbf{Key Findings.}
This case study demonstrates three practical properties of \model beyond the controlled benchmarks in prior subsections. \textit{First}, the method generalizes to open-ended, real-world agent tasks spanning 16 qualitatively distinct categories---not just math, coding, and tool-use. \textit{Second}, the cost advantage is realized under actual API pricing: the student+\model combination delivers teacher-surpassing quality at roughly two-thirds the cost of the teacher alone. \textit{Third}, the skill injection overhead is negligible at test time: on the 23 ``easy'' tasks where all three configurations pass, \model costs \$0.01 per task---identical to the bare student---indicating that retrieved skills do not inflate token consumption on tasks the student already handles well. These findings corroborate our earlier analyses: the Ralph Loop's student-side validation ensures stored knowledge is empirically effective, and student-aware synthesis prevents the teacher from producing guidance the weak executor cannot follow.

%% file: relate.tex
\section{Related Work}
\label{app:relate}

\subsection{Self-Improving Agents from Environment Feedback}
LLM agents interleave language reasoning with environment actions (\textit{e.g.}, ReAct~\cite{yao2023react} and Toolformer-style tool-use~\cite{schick2023toolformer}). Beyond one-shot prompting, a self-evolving agent literature studies how agents improve from interaction rather than offline expert labels, often leveraging sparse environment feedback as the main supervision for accumulating reusable competence~\cite{gao2025survey}. Representative approaches differ in what they evolve and what feedback they require: verbal self-improvement systems treat language as mutable memory by writing critiques after failures (\textit{e.g.}, Reflexion~\cite{shinn2023reflexion} and Self-Refine~\cite{madaan2023self}), while experience-driven skill libraries extract reusable procedures from interaction to grow pools or code-based repositories (\textit{e.g.}, ExpeL~\cite{zhao2024expel} and Voyager~\cite{wang2023voyager}). More modular self-evolution and long-horizon memory frameworks push adaptation toward reusable protocols or persistent external stores (\textit{e.g.}, Alita~\cite{qiu2025alita} and OMNE~\cite{jiang2024long}), and playbook-style context engineering closes execution-reflection-curation loops by revising structured bullets (\textit{e.g.}, ACE~\cite{zhang2025agentic}). Overall, most self-evolving agents assume same-capability evolution and adapt online at test time, learning from successes or rich feedback, whereas \model instead performs offline "brewing" of environment-validated, student-aware notes under binary supervision only, and then serves them via a frozen memory at test time.

\subsection{Knowledge Transfer and Lifelong Memory for LLM Agents}
Knowledge distillation has been widely used to compress large language models and has recently been extended to agent settings by transferring behavior from a teacher to a student across different transfer channels~\cite{hinton2015distilling,shridhar2023distilling}. At the trajectory level, structured span supervision and retrieval-augmented rollouts train students to imitate interleaved reasoning-action traces~\cite{liu2025structured,kang2026distilling}. In contrast, structure-level distillation compresses trajectories into interaction graphs or subgoal plans to preserve high-level task structure~\cite{chen2024magdi,hashemzadeh2024sub}. Training-free approaches further store reusable external artifacts: for example, AgentDistill extracts Model-Context-Protocols (MCPs) from successful teacher trajectories into an inference-time MCP-Box~\cite{qiu2025agentdistill}. RAG Distillation similarly distill teacher-to-student knowledge from failure-extracted hints, but internalize it into model weights via fine-tuning at the teacher's abstraction level, without calibrating knowledge to the student's capability or supporting incremental accumulation without re-training~\cite{ibrahim2025fine}. Related memory-based and lifelong-retrieval settings maintain external knowledge without parameter updates~\cite{jiang2024long,packer2023memgpt,zhao2026ama,trivedi2024appworld}.Overall, our work addresses the limitations of protocol-like artifact transfer and teacher-dependent abstractions by generating student-validated corrective notes from student failures using only binary rewards.

%% file: conclusion.tex
\section{Conclusion}
\label{sec:conclusion}

We study a practical deployment scenario for LLM agents where a strong teacher is available during training but only a fixed, weaker student serves at test time---with no weight updates, no teacher access, and only binary pass/fail feedback. The core challenge is that teacher knowledge, however insightful, often fails when executed by a substantially weaker student. We introduced \model, a training-free brew--serve framework built on the principle that knowledge must be calibrated to its executor. Its three components directly address the key obstacles: a failure-triggered teacher with Ralph Loop validation converts sparse binary feedback into environment-certified notes (Q1); student-aware synthesis paired with reactive validation bridges the teacher--student capability gap (Q2); and an append-only skill-scoped memory accumulates knowledge across tasks without modifying the student model (Q3). Extensive experiments across coding, math, tool-use, and a real-world Terminal-Bench deployment demonstrate that \model consistently lifts a fixed weak student, surpassing teacher-agnostic memory baselines.

%% file: appendix.tex
\appendix
\clearpage
\section{Appendix / supplemental material}

\subsection{Test-Time Serving Process.}
\label{app:test_time}
\begin{algorithm}[h]
\caption{Test-time serving on $\mathcal{D}_{\mathrm{test}}$ with frozen $\mathcal{M}^\star$.}
\label{alg:serving}
\small
\SetAlgoLined
\DontPrintSemicolon
\SetSideCommentLeft
\SetFillComment
\KwIn{$\mathcal{D}_{\mathrm{test}}$, frozen memory $\mathcal{M}^\star$, student agent $\mathcal{S}$, environment $\mathrm{Env}$, top-$k$ $k$, rerank params $\rho$, $\phi$}
\For{each task $x \in \mathcal{D}_{\mathrm{test}}$}{
    $q \gets \textsc{EncodeQuery}(x)$\;
    $\mathcal{M}_{\mathrm{scope}} \gets \textsc{RouteScope}(x,\, \mathcal{M}^\star)$\;
    $\mathcal{C} \gets \textsc{Retrieve}(\mathcal{M}_{\mathrm{scope}},\, q)$\;
    $\mathcal{N} \gets \textsc{Rerank}(\mathcal{C},\, q,\, \rho,\, \phi)$\tcp*[t]{Eq.~\ref{eq:note-rank}; soft suppression}
    $\mathcal{N}_k \gets \textsc{TopK}(\mathcal{N},\, k)$\;
    \If{$\mathcal{N}_k = \emptyset$}{
        $\tau \gets \textsc{Rollout}(\mathcal{S},\, x,\, \emptyset)$\tcp*[t]{vanilla fallback}
    }
    \Else{
        $\tau \gets \textsc{Rollout}(\mathcal{S},\, x,\, \mathcal{N}_k)$\;
    }
    $R \gets \textsc{Eval}(\mathrm{Env},\, x,\, \tau)$\;
}
\end{algorithm}

Algo.~\ref{alg:serving} formalizes the test-time inference procedure introduced in Sec.~\ref{sec:test_time_inference} and Sec.~\ref{sec:framework_overview}. For each test task $x$, the instruction is first encoded into a retrieval query $q$ that captures key textual signals such as task descriptions and tool names (line~2). The skill-scoped router $\textsc{RouteScope}$ then selects a partition $\mathcal{M}_{\mathrm{scope}} \subseteq \mathcal{M}^\star$ following the scoping policy in Sec.~\ref{sec:memory_architecture}---explicit scope hints in $x$ are matched first, then lexical similarity against category descriptions, and finally runtime cues when available---confining subsequent retrieval to a relevant skill subset and preventing cross-domain interference (line~3). Candidate notes $\mathcal{C}$ are drawn from $\mathcal{M}_{\mathrm{scope}}$ via dense or hybrid similarity scoring and reranked via Eq.~\ref{eq:note-rank}, which modulates each note's retrieval similarity by its historical \texttt{quality\_score} $\rho(n)$ and a frequency penalty $\phi(n)$; soft suppression dampens notes below a relative-score floor without hard-dropping them, preserving recall when the candidate pool is sparse. The top-$k$ notes $\mathcal{N}_k$ are then formatted and prepended to $\mathcal{S}$'s system prompt (lines~4--5). If $\mathcal{N}_k$ is non-empty, $\mathcal{S}$ performs a single $\textsc{Rollout}$ with memory augmentation; if retrieval returns no notes, the system falls back to the student's vanilla policy with no augmentation (lines~6--8). Critically, no teacher agent, Ralph Loop validation, or provisional memory writes are invoked during serving---the frozen $\mathcal{M}^\star$ is consumed read-only, and every task is completed in exactly one rollout, matching the inference budget of a bare student. Finally, the environment returns a binary terminal reward $R \in \{0,1\}$ (line~9), consistent with the sparse-feedback regime used throughout brewing and requiring no test-time labels or oracle solutions.

\subsection{Detailed Setting of Benchmarks}
\label{app:benchmark}

\begin{table}[h]
\centering
\caption{Original benchmark statistics and our experimental splits.
``Brew'' is $\mathcal{D}_{\mathrm{train}}$ for note distillation; ``Test'' is $\mathcal{D}_{\mathrm{test}}$ for serving.}
\label{tab:dataset-stats}
\begin{tabular}{@{}llcccc@{}}
\toprule
\textbf{Scenario} & \textbf{Benchmark} &
\multicolumn{2}{c}{\textbf{Original}} &
\multicolumn{2}{c}{\textbf{Ours}} \\
\cmidrule(lr){3-4}\cmidrule(lr){5-6}
 & & \textbf{Train} & \textbf{Test} & \textbf{Brew} & \textbf{Test} \\
\midrule
Math & MATH & 7{,}500 & 5{,}000 & 1{,}000 & 5{,}000 \\
Math & GSM8K & 7{,}473 & 1{,}319 & 1{,}000 & 1{,}319 \\
\midrule
Coding & MBPP & $\sim$500 & $\sim$500 & 257 & 163 \\
Coding & MBPP+ & --- & 378 & --- & 154 \\
\midrule
Tool-use & AppWorld \texttt{train} & \multicolumn{2}{c}{90 (30 scen.)} & 90 & --- \\
Tool-use & AppWorld \texttt{test\_normal} & \multicolumn{2}{c}{168 (56 scen.)} & --- & 168 \\
Tool-use & AppWorld \texttt{test\_challenge} & \multicolumn{2}{c}{417 (139 scen.)} & --- & 417 \\
\bottomrule
\end{tabular}
\end{table}

Table~\ref{tab:dataset-stats} summarizes the original public benchmark sizes and the processed splits used in our experiments.

\paragraph{Math.}
MATH and GSM8K originally contain 7{,}500 and 7{,}473 training problems respectively.
For computational efficiency during brewing, we only use the first 1{,}000 problems from their official \texttt{train} splits.
For evaluation, we use the complete official \texttt{test} splits (5{,}000 for MATH and 1{,}319 for GSM8K), reporting accuracy under symbolic equivalence checking.

\paragraph{Coding.}
We build on the sanitized MBPP benchmark (427 Python tasks), which is a subset of the original MBPP corpus ($\sim$500/500 train/test partition).
Following the standard convention, we use tasks with $\mathrm{task\_id}\in[11,510]$ for \texttt{train} (257 tasks) and $\mathrm{task\_id}>510$ for \texttt{test} (163 tasks).
MBPP+ extends MBPP with stronger tests (378 tasks) but provides no explicit train/test split.
Thus, we construct a strictly held-out test set by filtering out all MBPP+ tasks whose IDs overlap with our MBPP \texttt{train} split, retaining 154 tasks for evaluation.
We report Pass@1 using native assertions for MBPP and EvalPlus tests for MBPP+.

\paragraph{Tool-use.}
AppWorld originally provides 729 interactive API tasks partitioned into \texttt{train} (90 tasks), \texttt{dev} (57), \texttt{test\_normal} (168), and \texttt{test\_challenge} (417).
We directly brew on the official \texttt{train} split and evaluate on the two official test splits of increasing difficulty: \texttt{test\_normal} and \texttt{test\_challenge}.
We report Task Goal Completion (TGC) over tasks and Scenario Goal Completion (SGC).